%% file: main.tex
\documentclass[10pt,twocolumn]{article}

\usepackage{amsmath}
\usepackage{amssymb}
\usepackage{booktabs}
\usepackage{balance}
\usepackage[font=small,labelfont=bf]{caption}
\usepackage{graphicx}
\usepackage{microtype}
\usepackage{needspace}
\usepackage{xcolor}
\usepackage{url}
\usepackage{hyperref}

\hypersetup{
    colorlinks=true,
    linkcolor=blue!55!black,
    citecolor=blue!55!black,
    urlcolor=blue!55!black,
    pdftitle={Beyond Dense Adam States: Adaptive Log-Space Quantization for Memory-Efficient Optimizers},
    pdfauthor={Yan Wang}
}

\newcommand{\AL}{\textsc{AL}}
\newcommand{\ALviii}{\textsc{AL8}}
\newcommand{\ALxvi}{\textsc{AL16}}
\newcommand{\UFviii}{\textsc{UF8}}

\title{Beyond Dense Adam States:\\
Adaptive Log-Space Quantization for Memory-Efficient Optimizers}
\author{Yan Wang\\
\small Independent Researcher\\[0.35em]
\small
\raisebox{-0.15ex}{\includegraphics[height=1.25ex]{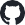}}\,
\href{https://github.com/yanfeiwong/adafactor-8bit}{\texttt{Implementation}}
\enspace$\cdot$\enspace
\href{https://github.com/yanfeiwong/al-quantization}{\texttt{Artifacts}}}
\date{}

\begin{document}
\maketitle

\begin{abstract}
Optimizer-state quantization is commonly designed for Adam's dense,
parameter-aligned first- and second-moment arrays. This abstraction breaks for
memory-efficient optimizers, whose states may be factored,
confidence-modulated, or maintained in a projected space, so similar
reconstruction error can produce different update error. We formulate
optimizer-state quantization as a joint problem over representation, topology,
and update semantics. We then introduce Adaptive Log-Space (\AL) quantization
for non-negative states. \AL{} fits each block's observed nonzero logarithmic
interval and reserves a separate code for exact zero, enforcing
$q=0\Leftrightarrow x=0$; signed momentum and state precision remain
independently selectable. Controlled probes show that adaptive ranges reduce
update error and temporal drift, exact-zero reservation preserves dormant
states, and state topology constrains useful block granularity. End-to-end
language-model training evaluates the resulting policy across dense, factored,
confidence, and projected optimizer states. On TinyLlama-1.1B, \ALviii{} with
uniform 8-bit momentum reaches 72.90 perplexity versus 73.54 for bitsandbytes
8-bit AdamW, with comparable optimizer-state storage and higher throughput.
CAME matches reference-level final perplexity across three seeds when its
non-negative states use \ALxvi{}, while a semantic grouping-and-protection
policy closes most of quantized Adafactor's 100K-step late-loss gap. These results make
state topology and update semantics first-class design constraints for
optimizer quantization.
\end{abstract}

\input{sections/introduction}
\input{sections/related_work}
\input{sections/state_analysis}
\input{sections/method}
\input{sections/controlled_analysis}
\input{sections/experiments}
\input{sections/implementation}
\input{sections/conclusion}

\bibliographystyle{plain}
\begingroup
\footnotesize
\setlength{\baselineskip}{9pt}
\bibliography{references}
\endgroup

\clearpage
\appendix
\input{sections/appendix}
\end{document}

%% file: sections/introduction.tex
\section{Introduction}
\label{sec:introduction}

Optimizer-state quantization is usually posed as a codec problem: compress
Adam's dense first- and second-moment arrays \cite{adam,adamw} while preserving
the values they store. Block-wise 8-bit optimizers made this formulation
practical \cite{dettmers8bit}, and later work pushed the same dense states to
lower precision \cite{fourbit}. The formulation assumes that the state
topology and the operator consuming the reconstructed state remain fixed.
Memory-efficient optimizers violate both assumptions.

Adafactor replaces a matrix second moment with row and column factors
\cite{adafactor}; CAME adds factored confidence states that modulate momentum
\cite{came}; and APOLLO derives update scales from moments maintained in a
projected gradient space \cite{apollo}. Quantization error no longer follows
one element-wise path. A factor error can affect an entire reconstructed row or
column, confidence error feeds a residual-driven EMA, and projected-state error
perturbs a scale applied outside the stored space. Treating these states as
interchangeable flat arrays discards the structure that makes the optimizer
memory-efficient.

\begin{figure*}[t]
    \centering
    \includegraphics[width=\textwidth]{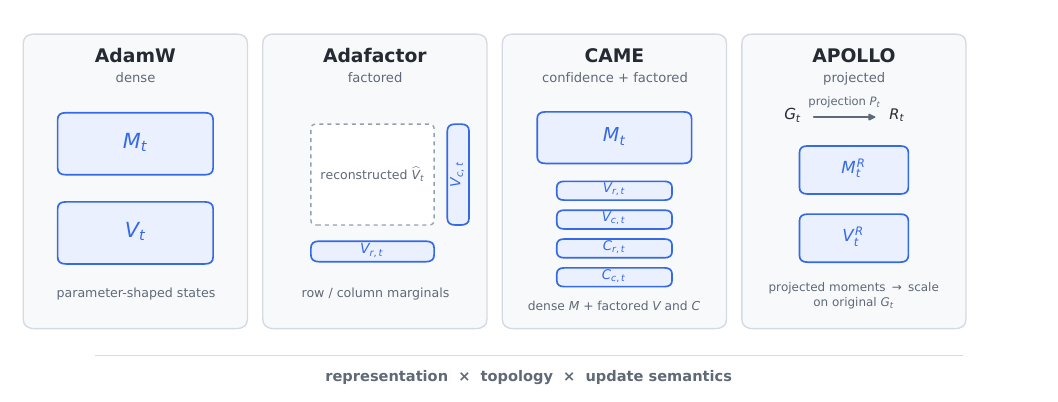}
    \caption{Optimizer states differ in topology and update semantics. AdamW
    stores dense parameter-shaped moments; Adafactor reconstructs a second-moment
    estimate from row and column marginals; CAME adds factored confidence states;
    and APOLLO maintains moments in a projected auxiliary space that determine a
    scale applied to the original gradient. State shapes are schematic.}
    \label{fig:state_landscape}
\end{figure*}

This paper places that structure inside the quantization problem. We separate
three interacting axes---state representation, state topology, and update
semantics---and evaluate fidelity after the optimizer-specific operator has
consumed the reconstructed state. State traces ground the distinction in actual
pre-training; controlled diagnostics isolate inverse-root sensitivity, factor
reconstruction, and confidence-state precision; and end-to-end runs test the
resulting policies across four optimizer paths, including projected-state
scaling.

Our representation is Adaptive Log-Space (\AL) quantization for non-negative
optimizer states. \AL{} directly maps each block's observed positive log
interval to an integer code space and reserves code zero exclusively for exact
zero, enforcing
$q=0\Leftrightarrow x=0$. Signed momentum uses an independent encoding, while
precision and grouping remain selectable by state role and topology. This
separates the representation of a non-negative block from the policy that
deploys it inside an optimizer.

We make three contributions:

\begin{itemize}
    \item We formulate optimizer-state quantization as a joint problem over
    representation, topology, and update semantics, and show how dense,
    factored, confidence, and projected update paths consume reconstruction
    error differently.
    \item We introduce \AL{}, a direct integer log-space representation that
    fits each block's observed positive interval and enforces the bidirectional
    exact-zero invariant. Signed momentum, state precision, and grouping remain
    independent policy decisions.
    \item Controlled update-error and drift diagnostics, including a
    matched-width comparison with SOLO's adaptive QEMA8 representation,
    identify the decisions tested in 20K-step TinyLlama-1.1B and 100K-step
    GPT-2 training. The evaluation shows that \ALviii{}+\UFviii{} improves over
    bitsandbytes (bnb) 8-bit AdamW, while CAME and Adafactor require precision
    and grouping policies matched to their structured states.
\end{itemize}

The resulting policy follows the optimizer's structure. \ALviii{} is effective for
dense AdamW second moments; CAME assigns higher precision to sensitive
non-negative states; Adafactor benefits from finer blocks and semantic
grouping with parameter protection; and APOLLO tests applicability on a
projected-state path. The common principle is to quantize the state an optimizer
actually maintains according to the operator that consumes it.

%% file: sections/related_work.tex
\section{Background and Related Work}
\label{sec:related_work}

\subsection{Quantized Optimizer States}

Dettmers et al. introduced practical 8-bit optimizers based on block-wise
nonlinear quantization \cite{dettmers8bit}. Independent blocks limit outlier
influence, while a fixed nonlinear codebook allocates resolution across state
magnitudes. We evaluate its bitsandbytes (bnb) implementation as the principal
8-bit AdamW baseline. Li et al. subsequently identified a second-moment
zero-point failure: a small positive value mapped to zero can be amplified by
the inverse-square-root preconditioner \cite{fourbit}. Their 4-bit method
excludes that zero point and combines small blocks with row- and column-wise
information.

SOLO adapts a geometric log codebook per block from its maximum and a percentile
statistic, targeting 2--4-bit EMA states \cite{solo}; its implementation also
provides an 8-bit QEMA variant. QEMA anchors its smallest reconstruction level
at the block percentile, whereas \AL{} maps the observed positive minimum and
maximum directly to a uniform integer grid in log space and reserves a distinct
code for exact zero. Section~\ref{sec:adaptive_log_comparison} compares the two
adaptive log constructions at matched width and block metadata.
COAT also adapts to observed group range, but through a value-space power
transform followed by an FP8 exponent--mantissa grid \cite{coat}; it is not a
direct integer log-space codec. These methods establish adaptive range use as
an important principle. Our scope joins the representation to native factored,
confidence, and projected state topologies and evaluates fidelity after their
consuming operators.

Recent analysis models stalling in quantized EMA states and the effect of state
resets \cite{quantstaleness}; STQuant dynamically allocates precision across
layers, states, and training steps \cite{stquant}. We use temporal drift to
compare representations, fix precision by optimizer-state role, and test those
policies across structured update paths.

\subsection{Memory-Efficient Optimizer Structure}

Adafactor reduces the quadratic state associated with a matrix parameter by
factorizing its second moment into row and column statistics \cite{adafactor}.
CAME augments this factored preconditioner with confidence-guided momentum,
introducing additional factored residual statistics \cite{came}.  These
algorithms reduce state size before quantization, but factorization also changes
how an error in one stored value propagates across a reconstructed matrix.

Low-rank gradient methods provide another route to optimizer-state reduction.
GaLore performs full-parameter learning while maintaining optimizer statistics
for gradients projected into a low-rank subspace \cite{galore}.  APOLLO further
uses moments in a projected space to derive channel- or tensor-wise scaling of
the original gradient \cite{apollo}.  We use APOLLO as a projected-state case:
the purpose is to test whether the same quantization machinery remains useful
when stored states determine a derived scaling operator applied outside the
stored space.

Prior work spans quantization precision, state factorization, and gradient
projection. Our point of departure is to analyze these axes jointly, holding
optimizer topology fixed while asking how its geometry and state semantics
determine quantized representation and grouping.

%% file: sections/state_analysis.tex
\section{Optimizer States Beyond Dense Adam}
\label{sec:structured_states}

Most optimizer-state quantization methods inherit a dense-Adam abstraction:
each state is treated as a parameter-aligned array, and the encoding is judged
by how accurately that array is reconstructed. Memory-efficient optimizers
break this abstraction in two independent ways. They change the topology of
the stored state, and they change the operator through which that state affects
the parameter update. We therefore separate three design axes throughout this
work: \emph{state representation}, \emph{state topology}, and \emph{update
semantics}. Figure~\ref{fig:state_landscape} illustrates the distinction.

\subsection{Topology Determines the Error Path}
\label{sec:state_topology}

Consider a matrix parameter $W \in \mathbb{R}^{m \times n}$ with gradient
$G_t$. A conventional Adam-style optimizer maintains dense first- and
second-moment states, $M_t,V_t \in \mathbb{R}^{m \times n}$. Ignoring bias
correction for clarity, its adaptive update is
\begin{equation}
    U_t = \frac{M_t}{\sqrt{V_t} + \epsilon}.
    \label{eq:adam_update}
\end{equation}
Here the stored states, preconditioner, and parameter update are aligned
element-wise. Quantization error therefore enters through a local inverse-root
rescaling of the corresponding momentum element.

Adafactor changes the error path by replacing a matrix second moment with row
and column statistics, $V_t^{(r)} \in \mathbb{R}^{m}$ and
$V_t^{(c)} \in \mathbb{R}^{n}$, from which it reconstructs an approximate
element-wise preconditioner \cite{adafactor}. One perturbed stored statistic
can consequently affect an entire row or column of the update. The state is
smaller precisely because its topology is no longer parameter-aligned.

CAME separates roles that are easy to conflate at the representation level. A
factored matrix path maintains dense momentum $M_t$, factored second moments
$V_t^{(r)},V_t^{(c)}$, and factored confidence states
$C_t^{(r)},C_t^{(c)}$ \cite{came}. Both $V$ and $C$ are non-negative
exponential moving averages, but $V$ constructs the adaptive denominator while
$C$ tracks a residual and modulates momentum. Similar storage types therefore
need not have similar precision requirements.

APOLLO introduces a complementary projected-state path \cite{apollo}. Its
first and second moments, $M_t^{(P)}$ and $V_t^{(P)}$, live in a low-rank
gradient space. APOLLO uses these projected moments to derive a channel-wise or
tensor-wise scale that is applied to the original gradient. Quantization error
thus perturbs a derived scaling operator outside the stored space.

The distinction goes beyond tensor shape. Dense, factored, confidence, and
projected states define different routes from stored error to parameter error.
A representation can be held fixed while topology changes which elements share
an encoding block and update semantics changes how the reconstructed values are
consumed. Quantization quality must therefore be evaluated after both choices,
not inferred from the codec alone.

\subsection{Fidelity in the Update Domain}
\label{sec:update_error}

Training does not consume an optimizer state directly; it consumes the update
produced from that state. A reconstruction metric such as
$\|S_t-\widehat{S}_t\|_2$ characterizes the representation, but state
reconstruction error alone does not determine update fidelity.

Let $\mathcal{U}(S_t,G_t)$ denote the optimizer-specific operator that maps a
state tuple $S_t$ and gradient $G_t$ to an update. We define relative update
error as
\begin{equation}
    \mathcal{E}_{\mathrm{upd}} =
    \frac{\left\|\mathcal{U}(\widehat{S}_t,G_t)-
    \mathcal{U}(S_t,G_t)\right\|_2}
    {\left\|\mathcal{U}(S_t,G_t)\right\|_2+\varepsilon},
    \label{eq:update_error}
\end{equation}
where $\varepsilon$ avoids instability when the reference update norm is close
to zero.

Equation~\eqref{eq:update_error} exposes effects hidden by state-space error. An
inverse square root can amplify a small perturbation; a factored statistic can
spread error across reconstructed elements; confidence error can feed back
through a residual EMA; and projected-state error can alter a scale applied to
the original gradient. Conversely, state error in an insensitive region of an
operator may have little effect on its output.

Update error is a controlled diagnostic, not a substitute for training. We use
it together with state reconstruction, temporal drift, loss trajectories, and
final evaluation metrics. Its role is to provide a common fidelity measure
after optimizer-specific topology and semantics have taken effect.

\begin{figure*}[t]
    \centering
    \includegraphics[width=\textwidth]{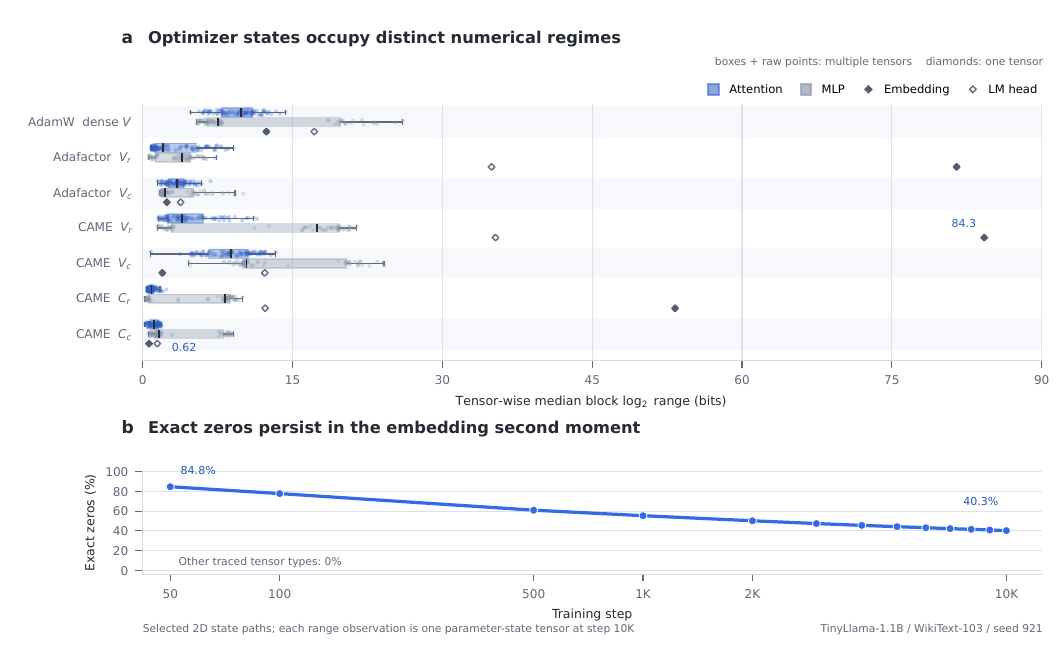}
    \caption{Optimizer states occupy distinct numerical regimes in a
    TinyLlama-1.1B trace. (a) Tensor-wise median block $\log_2$ ranges at step
    10K for selected two-dimensional state paths. Boxes summarize multiple
    attention or MLP parameter-state tensors and retain the raw observations;
    diamonds denote singleton embedding and output-head tensors. All values use
    one common linear bit-range axis. (b) The exact-zero fraction of the AdamW
    embedding second moment decreases from 84.8\% at step 50 to 40.3\% at step
    10K, while the other traced tensor types remain dense. The panels report
    summaries of recorded state traces.}
    \label{fig:state_heterogeneity}
\end{figure*}

\subsection{Observed States Are Heterogeneous}
\label{sec:state_heterogeneity}

The structural distinction matters because real optimizer states occupy
multiple numerical regimes. We inspect traces collected during TinyLlama-1.1B
pre-training on WikiText-103. Unless otherwise specified, the statistics use
batch size 4, sequence length 512, a nominal block size of 2048, and step 10K.
For each block, we define its bit range as
$\max_{x_i>0}\log_2x_i-\min_{x_i>0}\log_2x_i$. For each parameter-state
component, we compute the median block range; parameter-type summaries use a
state-size-weighted mean of these component medians.

In a factored Adafactor run, the state-size-weighted summaries at step 10K are
76.7 bits for embedding second-moment statistics, compared with 3.5 for
attention, 4.3 for MLP, and 2.4 for normalization parameters; the output-head
summary is 33.0. The embedding row statistics are particularly extreme, with
a median block range of 81.5 bits.

Exact zeros form a separate axis. In a full-rank AdamW trace, 84.8\% of the
embedding second-moment state is exactly zero at step 50, and 40.3\% remains
zero at step 10K, while traced attention, MLP, normalization, and output-head
states are dense. Numerical range, exact zero, and tensor topology therefore
define three separate dimensions of quantization difficulty.

CAME further separates state type from tensor type. Its embedding row
second-moment statistic exhibits a median block range of 84.3 bits, while the
embedding confidence row reaches 53.3 bits. Most attention confidence vectors
are much narrower, yet confidence enters a residual-driven EMA and subsequently
modulates momentum. Range alone cannot determine a safe precision; temporal
feedback and the downstream operator must also be tested.

Together, the traces impose three requirements. The represented nonzero range
must adapt to local state statistics; exact zero must remain distinct from a
small positive value; and precision and grouping must remain selectable by
state role and topology. Section~\ref{sec:adaptive_log} gives the
representation-level response, while Sections~\ref{sec:controlled_analysis}
and~\ref{sec:llm_evaluation} test the resulting decisions.

%% file: sections/method.tex
\section{Adaptive Log-Space Quantization}
\label{sec:adaptive_log}

\begin{figure*}[t]
    \centering
    \includegraphics[width=\textwidth]{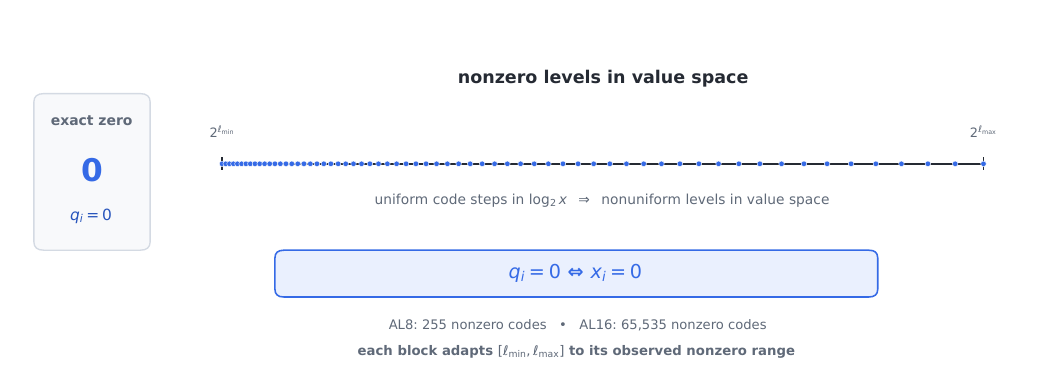}
    \caption{Adaptive Log-Space quantization reserves code zero exactly and
    places positive reconstruction levels uniformly in log space over a
    block-local observed interval. Marker density is schematic; the stated
    \ALviii{} and \ALxvi{} code counts are exact.}
    \label{fig:al_pipeline}
\end{figure*}

Section~\ref{sec:structured_states} identifies three design requirements:
adapt the represented nonzero range to local state statistics, preserve exact
zero as a distinct state, and allow numerical policy to follow state role and
topology. Adaptive Log-Space (\AL) quantization addresses the first two at the
representation level. It directly maps a block-local observed positive log
interval to integer codes and separates exact zero from every positive value.
Independent signed-momentum encoding and state-specific precision address the
third requirement at the optimizer-policy level.
Figure~\ref{fig:al_pipeline} summarizes the encoding.

\subsection{Adaptive Nonzero Log-Space Encoding}
\label{sec:al_encoding}

Consider a block $x\in\mathbb{R}_{\geq0}^{B}$ with block size $B$. For its
nonzero elements, we define the logarithmic interval
\begin{equation}
\begin{aligned}
    \ell_{\min} &= \max\!\left(\ell_{\mathrm{floor}},
        \min_{i:x_i>0}\log_2x_i\right), \\
    \ell_{\max} &= \max_{i:x_i>0}\log_2x_i .
\end{aligned}
\label{eq:al_bounds}
\end{equation}
Here, $\ell_{\mathrm{floor}}$ is a numerical lower bound. A fixed global log
range spends the same code space on every block; \AL{} fits the represented
nonzero interval independently to each block.

Writing $\Delta_{\ell}=\ell_{\max}-\ell_{\min}$, we normalize each positive
value as
$z_i=\operatorname{clip}((\log_2x_i-\ell_{\min})/\Delta_{\ell},0,1)$. Let
$L$ denote the total number of codes. Code zero is reserved exclusively for
exact zero, and positive values use the remaining $L-1$ codes:
\begin{equation}
q_i =
\begin{cases}
    0, & x_i=0, \\[1mm]
    1+\operatorname{round}\!\left[(L-2)z_i\right], & x_i>0 .
\end{cases}
\label{eq:al_quant}
\end{equation}

Dequantization maps nonzero codes uniformly across the stored log interval:
\begin{equation}
\widehat{x}_i =
\begin{cases}
    0, & q_i=0, \\[1mm]
    2^{\ell_{\min}+\frac{q_i-1}{L-2}\Delta_{\ell}}, & q_i>0 .
\end{cases}
\label{eq:al_dequant}
\end{equation}

We instantiate \AL{} at two precisions. \ALviii{} uses $L=256$, providing one
zero code and 255 positive reconstruction levels; \ALxvi{} uses
$L=65{,}536$, providing one zero code and 65,535 positive levels. Each block
also stores $\ell_{\min}$ and $\Delta_{\ell}$. An all-zero block retains only
zero codes.

The implementation handles degenerate ranges without changing these semantics.
If all positive values coincide, it expands the stored log interval by one unit
and clamps its width to at least $10^{-12}$. The CUDA path caps
$\ell_{\max}$ at 126, below FP32 exponentiation overflow. The optimizer path
supplies $\ell_{\mathrm{floor}}$, typically as the log of the squared numerical
floor used in its second-moment update.

\subsection{Exact Zero Is a Semantic State}
\label{sec:exact_zero}

Exact zero is not interchangeable with the smallest positive reconstruction
level. A dormant optimizer entry can remain zero until its parameter first
receives a gradient; assigning it a positive value creates state history that
the reference optimizer does not contain. Conversely, mapping a small positive
second moment to zero can make an inverse-root update unstable.

\AL{} assigns disjoint code spaces to these cases. Equations~\eqref{eq:al_quant}
and~\eqref{eq:al_dequant} enforce
\begin{equation}
    \boxed{q_i=0 \iff x_i=0}.
    \label{eq:exact_zero}
\end{equation}
Thus a positive state cannot collapse to exact zero, and a dormant state cannot
acquire an artificial positive value. Reserving one code reduces the positive
levels from 256 to 255 in \ALviii{}; Section~\ref{sec:controlled_analysis}
shows that this cost is negligible for dense states and consequential when
exact zeros are common.

This invariant complements the second-moment zero-point analysis of
Li et al.~\cite{fourbit}. Their failure case is a small positive value that
rounds to zero and is amplified by an inverse square root. Exact-zero
reservation additionally prevents a genuinely zero state from being
reconstructed as positive.

\subsection{Representation and State Policy}
\label{sec:state_precision}

\AL{} represents non-negative adaptive statistics. We apply it to second-moment
states $V$ and confidence states $C$, choosing \ALviii{}, \ALxvi{}, or FP32
according to state sensitivity and memory cost. Signed first moments $M$ use an
independent encoding because their distribution and update role differ.

For signed momentum, controlled traces select uniform 8-bit quantization
(\UFviii{}): at equal bit width it gives lower relative $L_2$ error than the
evaluated Dettmers-style fixed nonlinear codebook at both measured training
steps, with no observed sign flips. \UFviii{} supplies the signed-momentum
component of the end-to-end policy, while \AL{} defines its non-negative-state
representation. Semantic grouping separately determines which states are
encoded and which remain unquantized.

Separating representation from policy is essential for structured optimizers.
A dense second moment may justify \ALviii{} because state bytes dominate; a
small but update-sensitive confidence state can receive \ALxvi{} or FP32 at
little absolute cost. Section~\ref{sec:llm_evaluation} tests this allocation in
CAME and tests semantic grouping and parameter protection in Adafactor.

\subsection{Block Granularity and Memory Cost}
\label{sec:block_granularity}

Smaller blocks localize the fitted interval, while larger blocks amortize its
metadata. Each block stores two FP32 values, $\ell_{\min}$ and
$\Delta_{\ell}$. Ignoring tensor-boundary padding, the approximate storage per
state element is
\begin{equation}
    c_{\mathrm{AL8}}\simeq1+\frac{8}{B}, \qquad
    c_{\mathrm{AL16}}\simeq2+\frac{8}{B},
    \label{eq:al_memory}
\end{equation}
where the first term is the code in bytes and the second is shared FP32
metadata. Section~\ref{sec:controlled_analysis} evaluates the resulting
localization--metadata trade-off.

At the $B=2048$ granularity used by the core AdamW configuration, the two
endpoints cost $8/2048=0.0039$ bytes per state element. The evaluated bnb
baseline uses 256-element state blocks with one FP32 scale, or
$4/256=0.0156$ metadata bytes per element \cite{dettmers8bit}. Thus \ALviii{}
stores one additional
scalar per block but incurs one quarter of the per-element block metadata at
the tested granularity. Its encoder obtains the two endpoints by block
reductions and then applies the closed-form log-affine map in
Eq.~\eqref{eq:al_quant}; decoding applies Eq.~\eqref{eq:al_dequant} directly.
Section~\ref{sec:llm_evaluation} reports the resulting system-level storage and
throughput with the fused implementation.

\ALxvi{} codes occupy a 16-bit integer container and are interpreted as
unsigned values during quantization and dequantization. Metadata remains FP32
for both precisions; Eq.~\eqref{eq:al_memory} describes the logical code space
independently of the host container's signedness.

%% file: sections/controlled_analysis.tex
\section{Controlled Tests of the Design Claims}
\label{sec:controlled_analysis}

The method makes four testable claims: local adaptation should remove the
global-floor trade-off; exact-zero reservation should preserve dormant state;
update-domain fidelity should persist through temporal accumulation; and
precision and block topology should remain independent decisions. End-to-end
loss cannot isolate these mechanisms, so we test them under controlled
distributions before evaluating language-model training.

\begin{figure*}[t]
    \centering
    \includegraphics[width=\textwidth]{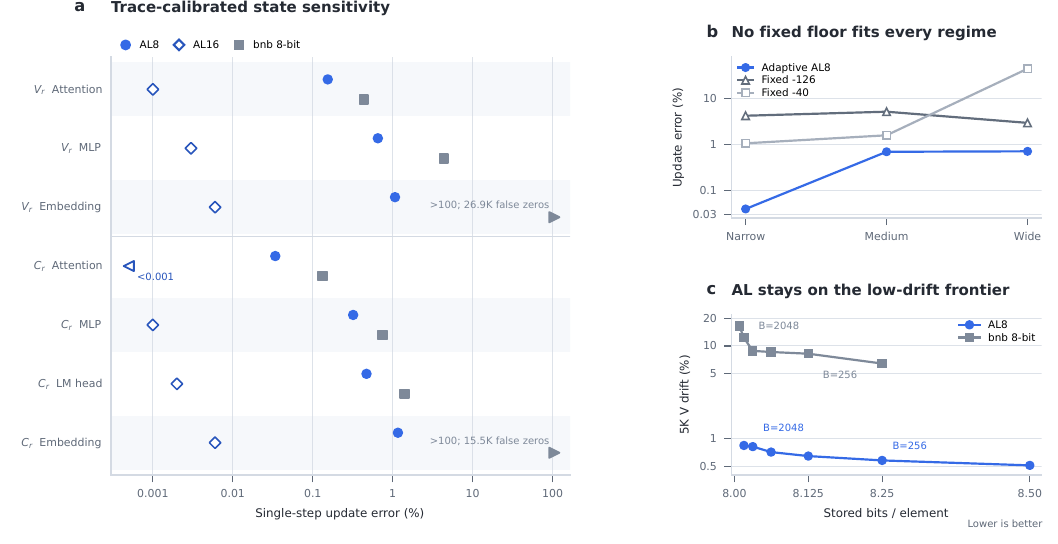}
    \caption{Controlled diagnostics separate state-specific sensitivity,
    range adaptation, and temporal drift. (a) Single-step update error for
    synthetic blocks calibrated to the step-10K CAME trace by tensor-wise
    median and median block range. $V_r$ and $C_r$ denote row factors of the
    factored second-moment and residual-confidence states. Hollow left triangles
    mark \ALxvi{} values below 0.001\%; filled right triangles mark bnb 8-bit values
    above 100\%, whose labels also report nonzero values mapped to zero. (b) A
    block-adaptive interval avoids the failure modes of two fixed logarithmic
    ranges across narrow, medium, and wide synthetic regimes. (c) In a 5K-step
    second-moment EMA, \ALviii{} remains on a lower drift--storage frontier than
    the source-aligned bnb 8-bit V-state reproduction. These panels isolate
    controlled mechanisms; Section~\ref{sec:llm_evaluation} reports end-to-end
    training quality.}
    \label{fig:controlled_fidelity}
\end{figure*}

All values come from the executed analysis notebook, with each comparison
sharing identical synthetic inputs across methods. Controlled bnb 8-bit
$V$-state experiments transcribe bitsandbytes' fixed unsigned nonlinear
codebook and per-block absmax quantize--dequantize path
\cite{dettmers8bit}; the end-to-end baseline in
Section~\ref{sec:llm_evaluation} runs the package directly.
Figure~\ref{fig:controlled_fidelity} summarizes the main results.

\subsection{Adaptive Ranges Eliminate Floor Selection}

Table~\ref{tab:adaptive_fixed} compares the relative update error from
Eq.~\eqref{eq:update_error} for a 2048-element block.  The tested distributions
are three log-normal regimes in base-2 space, denoted narrow, medium, and wide
according to their relative sampled log-spans. A single fixed logarithmic floor
cannot serve all three regimes. A high floor improves resolution for narrow
states but clips wide states; a very low floor spends most codes on an unused
range. Fitting each block's observed nonzero interval removes this global
choice and keeps update error below 1\% in all three cases.

\begin{table}[t]
\centering
\small
\caption{Single-step update error (\%, lower is better) for fixed and adaptive
logarithmic ranges, block size 2048.}
\label{tab:adaptive_fixed}
\resizebox{\columnwidth}{!}{%
\begin{tabular}{lrrrr}
\toprule
Distribution & Fixed $-126$ & Fixed $-53$ & Fixed $-40$ & Adaptive \\
\midrule
Narrow & 4.200 & 1.568 & 1.052 & \textbf{0.039} \\
Medium & 5.140 & 1.835 & 1.574 & \textbf{0.688} \\
Wide   & 2.934 & 1.993 & 44.505 & \textbf{0.704} \\
\bottomrule
\end{tabular}}
\end{table}

The global-floor mismatch compounds over time. In a 5K-step second-moment EMA with
new-observation weight $\alpha=0.001$ (equivalently, $\beta_2=0.999$) and block
size 2048, adaptive quantization produces 0.736\%
state drift.  Fixed floors at $-126$, $-53$, $-40$, and $-28$ produce 56.914\%,
52.956\%, 74.463\%, and 25.254\%, respectively; the source-aligned bnb 8-bit
reproduction produces 10.032\%. The result isolates the benefit of fitting the
represented log interval to each block and removes the need to select one
favorable global floor.

\subsection{Exact Zero Preserves Dormant State}

The invariant becomes consequential when true zeros are present. Reserving one
of 256 codes has almost no effect for dense blocks. For the
narrow dense distribution, the reserved and unreserved variants yield
0.079\% and 0.078\% state reconstruction error and the same 0.039\% update
error.  Their behavior diverges when true zeros are present.  In the 90\%-sparse
case in Table~\ref{tab:zero_ablation}, the no-zero variant reconstructs dormant
entries as positive values.  RMS clipping can hide their immediate contribution
to the measured update, but removing that clipping exposes an error exceeding
100\%.  The reserved design retains exact zeros and reduces the unclipped update
error to 0.122\%.

\begin{table}[t]
\centering
\small
\caption{Exact-zero ablation at block size 2048.  ``No clip'' removes the
optimizer clipping that otherwise masks dormant entries in this single-step
probe.}
\label{tab:zero_ablation}
\resizebox{\columnwidth}{!}{%
\begin{tabular}{llrr}
\toprule
Distribution & Codes & State error (\%) & No-clip update (\%) \\
\midrule
Dense, narrow & $1+255$ & 0.079 & 0.039 \\
Dense, narrow & $0+255$ & 0.078 & 0.039 \\
Sparse, 90\% & $1+255$ & 0.243 & 0.122 \\
Sparse, 90\% & $0+255$ & 8.963 & $>100$ \\
\bottomrule
\end{tabular}}
\end{table}

\Needspace{5\baselineskip}
The no-clipping probe isolates the artificial state introduced when exact zeros
are reconstructed as positive values. Section~\ref{sec:llm_evaluation} then
evaluates the complete representation policies end to end.

\subsection{Adaptive Log Codebooks Diverge After the Operator}
\label{sec:adaptive_log_comparison}

SOLO's QEMA is the closest prior adaptive log construction. We transcribe its
official 8-bit path at the pinned implementation commit recorded in the
artifact and retain its stochastic rounding. At the matched block size
$B=2048$, QEMA8 and \ALviii{} both use 8-bit codes and two FP32 metadata values
per block, or 8.03125 logical bits per element. Table~\ref{tab:qema_probe}
compares the representations before and after the consuming operator.

\begin{table}[t]
\centering
\small
\caption{Representation-level probes at matched 8-bit width and $B=2048$.
State error excludes true zeros. Operator error follows the consuming
inverse-square-root path; the sparse case uses the no-clipping diagnostic.
$0{\to}{+}$ counts true zeros reconstructed as positive.}
\label{tab:qema_probe}
\resizebox{\columnwidth}{!}{%
\begin{tabular}{llrrr}
\toprule
Probe & Codec & State error (\%) & Operator error (\%) & $0{\to}{+}$ \\
\midrule
Wide dense & \ALviii{} & 1.526 & 0.704 & -- \\
           & QEMA8     & \textbf{1.408} & 97.701 & -- \\
Sparse, 5\% & \ALviii{} & \textbf{0.262} & \textbf{0.131} & 0 \\
            & QEMA8     & 1.879 & 32.578 & 410 \\
Traced $V_r$ embedding & \ALviii{} & 1.107 & \textbf{1.069} & -- \\
                       & QEMA8     & \textbf{0.811} & 99.994 & -- \\
\bottomrule
\end{tabular}}
\end{table}

QEMA8 has lower state-space error in the wide dense and traced embedding
cases, yet its update error is roughly two orders of magnitude larger. The
percentile floor therefore changes the low-value tail that the inverse-square-root
operator amplifies. The sparse probe stays below QEMA's 10\% percentile
threshold, avoiding the degenerate $\alpha=0$ boundary; even there, all 410
true zeros reconstruct as positive. \ALviii{} preserves them and reduces
no-clipping update error from 32.578\% to 0.131\%. The artifact reports QEMA8's
default $B=128$ results and the $\alpha=0$ boundary separately. This comparison
isolates the representation; it is not an end-to-end reproduction of SOLO.

\subsection{Update Fidelity Persists Through Time}

For a 5K-step Adam-style EMA with FP32 momentum, \ALviii{} reduces both state
drift and steady-state update error relative to the bnb 8-bit reproduction.
At block size 256, \ALviii{} yields 0.562\% $V$ drift and 0.280\% update error,
versus 7.282\% and 3.436\% for the baseline.  At block size 2048, the
corresponding values are 0.736\% and 0.357\%, versus 10.032\% and 4.577\%.
The advantage persists in the update domain. Factored states can behave
differently: when a constructed row statistic spans
heterogeneous hot, normal, and cold groups, \ALviii{} has lower reconstructed
update error even when an $L_2$ state-drift metric favors the baseline on the
largest values. This disagreement directly demonstrates why the operator
output must be measured alongside the stored tensor.

\subsection{Topology Constrains Block Scale}

Block size couples topology to metadata. Smaller blocks improve localization
but add metadata. For a hypothetical
1.1B-element second-moment state, \ALviii{} uses 8.25 bits/element at $B=256$
and 8.031 bits/element at $B=2048$; the corresponding estimates are 1081.8 and
1053.1 MiB. In an independently generated 5K-step EMA Pareto sweep over block size and
representation, the measured drift rises from 0.577\% to 0.814\% between these
block sizes, while bnb 8-bit rises from 8.243\% to 12.259\%.  The
numerical gap favors larger blocks as a memory-efficient default in this
synthetic setting.

Topology places an additional constraint on that choice.  A 2048-element flat
block aligns naturally with the hidden width of the TinyLlama configuration,
but spans more than two 768-wide rows in GPT-2. The 100K experiment in
Section~\ref{sec:long_horizon} reports two observed contrasts: block size under
G0, and G1 protection at $B=256$.  The 20K Adafactor grid in
Table~\ref{tab:tinyllama_core} tests both grouping policies at both block sizes;
the 100K table isolates block size under G0 and grouping at $B=256$.

\subsection{Momentum Requires a Separate Encoding}

Signed first moments use a separate code space. At block size 256,
\UFviii{} produces 0.6954\% relative $L_2$ error on
the step-100 momentum sample, compared with 1.0711\% for the evaluated nonlinear
codebook; both have cosine similarity above 0.9999 and no measured sign flips.
At step 1000, the errors are 0.7126\% and 1.1134\%. The consistent advantage
selects the signed-momentum default used end to end.
Section~\ref{sec:llm_evaluation} then tests the resulting \UFviii{}+\AL{}
policies across dense, confidence, and projected optimizer paths.

\subsection{Precision Is a State Decision}

Precision should follow both sensitivity and state cardinality. Increasing a
dense 1.1B-element second moment from \ALviii{} to \ALxvi{} roughly
doubles its state storage, eliminating most of the memory benefit relative to
native FP16.  A factored confidence state is much smaller.  This asymmetry makes
mixed precision attractive: \ALviii{} can remain the default for large $V$
states while a sensitive, low-cardinality $C$ state receives \ALxvi{} or FP32.
The end-to-end CAME ablation in Table~\ref{tab:tinyllama_core} tests this policy.

%% file: sections/experiments.tex
\section{End-to-End Language-Model Evaluation}
\label{sec:llm_evaluation}

\subsection{Evaluation Design}

We evaluate randomly initialized causal language models on the
WikiText-103 raw corpus \cite{wikitext}.  The core table and trajectories use the
TinyLlama-1.1B architecture \cite{tinyllama} for 20K optimizer steps with
sequence length 512, per-device batch size 4, and seed 921.  Validation is run
every 1000 steps.  The long-horizon benchmark uses GPT-2 124M
\cite{gpt2} for 100K steps with sequence length 512 and batch size 16 under the
same initialization protocol; it records training metrics without periodic
validation. Models are initialized from configuration and trained from scratch
with BF16 computation, FP32 parameters, and gradient checkpointing.
The long-horizon run is used to compare stability and late training loss, not
held-out quality.

The learning-rate schedule is constant after 1000 warmup steps.  The effective
learning rate is $10^{-4}$ for AdamW, CAME, and APOLLO configurations and
$10^{-3}$ for Adafactor.  Weight decay and gradient clipping are disabled in
the reported presets.  Adam-style paths use $(\beta_1,\beta_2)=(0.9,0.999)$;
CAME additionally uses $\beta_3=0.9999$; APOLLO uses rank 256, channel-wise
scaling, and a projection update gap of 200.  The bnb 8-bit AdamW baseline
directly runs the bitsandbytes implementation corresponding to
Dettmers et al.~\cite{dettmers8bit}; it is not a reimplementation in our
codebase.  Its fixed nonlinear codebooks are paired with block-wise absmax
scales for the signed and non-negative states.  HF Adafactor, the official CAME
implementation, and the official APOLLO implementation provide the other
reference paths.

We distinguish two parameter-grouping policies. \textbf{G0} applies the
selected optimizer configuration without special protection, except where the
optimizer requires dedicated routing. \textbf{G1} separates one-dimensional
tensors, biases, normalization parameters, token embeddings, and the
language-model head; it leaves optimizer states for this group unquantized,
while the remaining tensors retain the selected low-precision path. G1 is a
predeclared coarse semantic grouping that tests parameter protection at the
aggregate group level.

The reported optimizer-state memory is the sum of CUDA tensor storage owned by the
optimizer.  Peak allocation is PyTorch's maximum allocated memory and excludes
allocator fragmentation; when a 20K run lacks the counter, the analysis report
uses a configuration-matched 1K proxy because the peak is reached early.
Throughput is measured in input tokens per second. The learning-rate,
batch-size, topology, and horizon sweeps test whether the main comparisons
persist under controlled changes.

\subsection{Core 20K-Step Results}

\begin{figure*}[t]
    \centering
    \includegraphics[width=\textwidth]{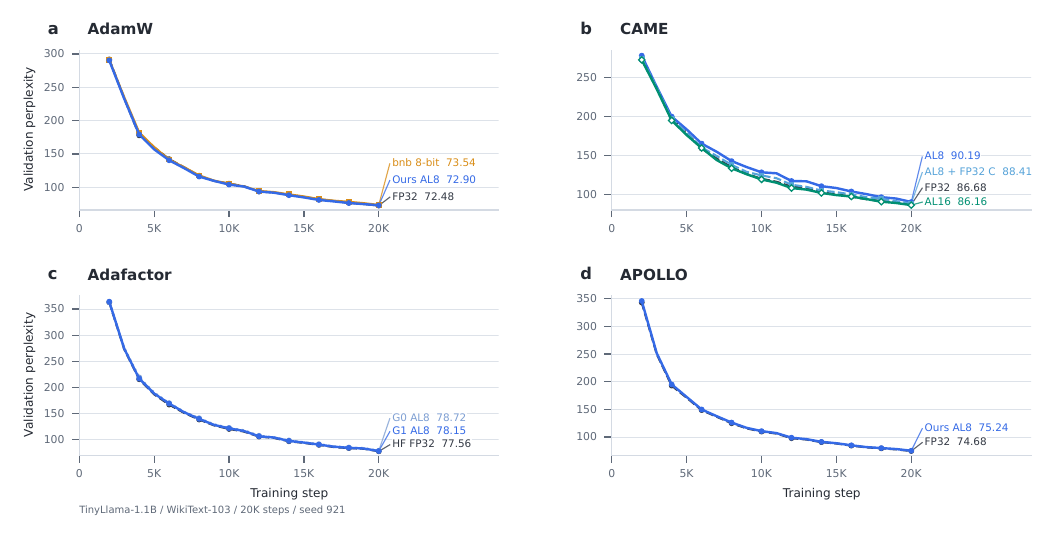}
    \caption{Validation perplexity across the 20K-step TinyLlama-1.1B
    benchmark for AdamW, CAME, Adafactor, and APOLLO. The horizontal axis begins
    at step 0, while the displayed curves begin at step 2K after omitting the
    1K evaluation for legibility; endpoint labels give the step-20K values in
    Table~\ref{tab:tinyllama_core}. Path-specific vertical axes are used for
    legibility. For Adafactor, G0 is
    the default grouping and G1 protects selected sensitive tensors as defined
    in the experimental setup.}
    \label{fig:training_curves}
\end{figure*}

Table~\ref{tab:tinyllama_core} tests the resulting policy across dense,
confidence, factored, and projected states; Figure~\ref{fig:training_curves}
shows the corresponding validation trajectories. For dense AdamW, \ALviii{}
second moments with \UFviii{} momentum
reaches 72.90 perplexity, a gap of 0.42 from FP32 and 0.64 lower than the
evaluated bnb 8-bit baseline. Despite storing two metadata values per
\ALviii{} block, the larger block granularity gives 2119.2 MiB of measured
optimizer state, slightly below bnb's 2131.8 MiB and far below FP32 AdamW's
8392.7 MiB. The fused implementation reaches 2960 tokens/s, compared with 2640
for bnb and 2502 for FP32. The peak-allocation reduction is smaller than the state reduction
because parameters, gradients, activations, and temporary buffers remain.

\begin{table*}[t]
\centering
\small
\caption{TinyLlama-1.1B, WikiText-103, 20K steps, batch size 4. Memory: MiB.
For the bnb 8-bit baseline, both $M$ and $V$ use bitsandbytes 8-bit encodings;
$B_V$ is the non-negative-state block size (momentum blocks: 256).}
\label{tab:tinyllama_core}
\resizebox{\textwidth}{!}{%
\begin{tabular}{lllccccrrrr}
\toprule
Path & Group & Configuration & $M$ & $V$ & $C$ & $B_V$ & PPL $\downarrow$ & Peak $\downarrow$ & State $\downarrow$ & tok/s $\uparrow$ \\
\midrule
AdamW & G0 & FP32 reference & FP32 & FP32 & -- & -- & 72.48 & 21046.0 & 8392.7 & 2502 \\
AdamW & G0 & bnb 8-bit & bnb & bnb & -- & 256 & 73.54 & 10608.6 & 2131.8 & 2640 \\
AdamW & G0 & Ours & \UFviii{} & \ALviii{} & -- & 2048 & 72.90 & 10596.8 & 2119.2 & 2960 \\
\midrule
CAME & G0 & FP32 reference & FP32 & FP32 & FP32 & -- & 86.68 & 13428.5 & 4203.2 & 1866 \\
CAME & G0 & Ours, all \ALviii{} & \UFviii{} & \ALviii{} & \ALviii{} & 2048 & 90.19 & 9509.0 & 1068.0 & 2432 \\
CAME & G0 & Ours, FP32 $C$ & \UFviii{} & \ALviii{} & FP32 & 2048 & 88.41 & 9510.0 & 1070.3 & 2423 \\
CAME & G0 & Ours, \ALxvi{} & \UFviii{} & \ALxvi{} & \ALxvi{} & 2048 & 86.16 & 9510.8 & 1069.8 & 2428 \\
\midrule
Adafactor & G0 & HF reference & -- & FP32 factored & -- & -- & 77.56 & 8934.7 & 3.6 & 2206 \\
Adafactor & G0 & Ours & -- & \ALviii{} factored & -- & 256 & 78.72 & 8442.6 & 1.2 & 2740 \\
Adafactor & G0 & Ours & -- & \ALviii{} factored & -- & 2048 & 79.36 & 8442.7 & 1.3 & 2734 \\
Adafactor & G1 & Ours & -- & mixed/\ALviii{} & -- & 256 & 78.15 & 8999.2 & 1.4 & 2708 \\
Adafactor & G1 & Ours & -- & mixed/\ALviii{} & -- & 2048 & 78.29 & 8999.3 & 1.5 & 2698 \\
\midrule
APOLLO & G0 & FP32 reference & FP32 & FP32 & -- & -- & 74.68 & 11560.2 & 2078.7 & 2331 \\
APOLLO & G0 & Ours & \UFviii{} & \ALviii{} & -- & 2048 & 75.24 & 10290.9 & 1694.7 & 2077 \\
\bottomrule
\end{tabular}}
\end{table*}

The CAME path tests whether one low precision can be assigned to every
non-negative state. The all-\ALviii{} configuration saves most optimizer-state
memory but has a 3.51
perplexity gap from the reference.  Keeping $C$ in FP32 reduces this gap to
1.73 with only 2.3 MiB additional measured state.  Using \ALxvi{} for both $V$
and $C$ reaches parity with the reference (86.16 versus 86.68) while preserving
the dominant saving from 8-bit momentum.  Together, these configurations
identify non-negative-state precision---not momentum storage---as the decisive
CAME variable in this setup.

Adafactor exposes topology effects even though its factored reference uses only
3.6 MiB of measured optimizer state. In the completed
$2\times2$ grid, $B=256$ is better than
$B=2048$ under both G0 (78.72 versus 79.36) and G1 (78.15 versus 78.29), while
G1 records lower final PPL at both block sizes. On the projected APOLLO path,
our configuration records a 0.56 PPL gap and 384.0 MiB lower measured state;
throughput is 2077 versus 2331 tokens/s for the reference in this
implementation.

\subsection{Robustness Across Seeds and Training Conditions}

In the three-seed AdamW and CAME comparisons, each quantized run is paired with
its same-seed FP32 reference. Mean $\pm$ SD PPL gaps are
$0.29\pm0.12$ for AdamW \UFviii{}+\ALviii{} and $0.69\pm0.41$ for bnb 8-bit.
The CAME precision separation is larger: all-\ALviii{} has a
$6.74\pm3.36$ gap, whereas assigning \ALxvi{} to its non-negative states
remains aligned with FP32 at $-0.15\pm0.40$.

The 10K learning-rate grid applies $\{0.1,1,10\}$ to a common $10^{-3}$ preset
at batch size 4; $0.1\times$ is nominal for AdamW, CAME, and APOLLO, and
$1\times$ for Adafactor. The grid reveals distinct sensitivity regimes across
optimizer paths. The
bnb 8-bit AdamW baseline collapses at $10\times$, while our \UFviii{}+
\ALviii{} AdamW remains finite; all-\ALviii{} CAME also collapses at $10\times$,
whereas the CAME reference, FP32-$C$ variant, and \ALxvi{} variant remain
finite.  Adafactor and APOLLO configurations remain finite across the three
tested multipliers.  Stability is therefore path-specific across the evaluated
configurations.

At the nominal learning rate, batches 4, 8, and 16 provide a second controlled
change.
AdamW \UFviii{}+\ALviii{} has trajectory mean absolute errors of 0.0078,
0.0053, and 0.0050 relative to FP32, compared with 0.0098, 0.0118, and 0.0111
for the bnb 8-bit baseline.  CAME's corresponding errors are 0.0495, 0.0603, and
0.0572; Adafactor's are 0.0266, 0.0249, and 0.0283; and APOLLO's are 0.0119,
0.0138, and 0.0148.  The batch sweep supports the AdamW result over the tested
range but also shows that the CAME precision gap is not an isolated final-step
fluctuation. Appendix~\ref{app:sensitivity_tables} reports the complete selected
learning-rate and batch-size sensitivity tables.

\subsection{Long-Horizon Topology Test}
\label{sec:long_horizon}

The 100K-step GPT-2 benchmark tests whether topology effects persist beyond the
20K-step evaluation envelope. Table~\ref{tab:gpt2_long} reports late training
loss. No listed configuration collapses. AdamW remains close to its
reference at both block sizes, with a smaller gap at $B=256$.  CAME and APOLLO
show the same block-size direction. Although $B=2048$ reduces metadata, the
consistent fidelity advantage of $B=256$ on the 768-wide GPT-2 configuration
is the topology constraint on that memory--error trade-off. The largest change
occurs for Adafactor:
G0 with $B=256$ has a late-loss gap of $+0.1185$, whereas G1 with the same block
size reduces the gap to $+0.0159$.  This is consistent with the hypothesis that
flat quantization blocks and uniform parameter treatment are poor matches for
sensitive tensors in the 768-wide GPT-2 topology.

\begin{table}[t]
\centering
\small
\caption{GPT-2 124M, 100K steps, batch size 16.  Late training loss and the
change from the path-specific full-precision reference.}
\label{tab:gpt2_long}
\resizebox{\columnwidth}{!}{%
\begin{tabular}{llrr}
\toprule
Path/configuration & Group & Late loss & $\Delta$ \\
\midrule
AdamW FP32 & G0 & 3.4564 & -- \\
AdamW bnb 8-bit, $B=256$ & G0 & 3.4648 & +0.0084 \\
AdamW ours, $B=256$ & G0 & 3.4603 & +0.0039 \\
AdamW ours, $B=2048$ & G0 & 3.4679 & +0.0115 \\
\midrule
CAME FP32 & G0 & 3.4271 & -- \\
CAME ours, $B=256$ & G0 & 3.4447 & +0.0176 \\
CAME ours, $B=2048$ & G0 & 3.4555 & +0.0284 \\
\midrule
Adafactor HF & G0 & 3.8522 & -- \\
Adafactor ours, $B=256$ & G0 & 3.9707 & +0.1185 \\
Adafactor ours, $B=2048$ & G0 & 4.0026 & +0.1504 \\
Adafactor ours, $B=256$ & G1 & 3.8681 & +0.0159 \\
\midrule
APOLLO FP32 & G0 & 3.8551 & -- \\
APOLLO ours, $B=256$ & G0 & 3.8683 & +0.0132 \\
APOLLO ours, $B=2048$ & G0 & 3.8789 & +0.0238 \\
\bottomrule
\end{tabular}}
\end{table}

At fixed $B=256$, G1 changes only the parameter-grouping policy and closes most
of the quantized Adafactor gap. Because G1 protects its tensor classes jointly,
the contrast identifies the aggregate grouping effect rather than an
individual protected class.

%% file: sections/implementation.tex
\section{Implementation and Reproducibility}
\label{sec:implementation}

\subsection{Reference and CUDA Paths}

One PyTorch optimizer interface exposes AdamW-style, Adafactor, CAME, and
APOLLO update paths. Non-negative states support \ALviii{}, \ALxvi{}, and FP32;
signed momentum supports uniform and nonlinear low-bit encodings or FP32, and
CAME confidence precision is independently configurable. A Python path defines
the update semantics. A JIT-compiled CUDA extension fuses dequantization, EMA
updates, requantization, and parameter updates where the selected path permits
it.

The frozen optimizer revision is Adafactor8Bit v0.4.3 at
\href{https://github.com/yanfeiwong/adafactor-8bit/commit/4f544d8b8eebaf50053a4e8a27096e79b049b480}{\texttt{4f544d8}}.
The public
\href{https://github.com/yanfeiwong/al-quantization}{\texttt{reproducibility artifact}}
maps each table and figure to its script and machine-readable inputs. Optimizer
checkpoints retain quantization type and metadata and support round-trip
save/load, resumption, and selected state-precision migration.

\subsection{Validation and Evaluation Scope}

Reference alignment compares the unquantized Python paths with the corresponding
public optimizer implementations; CUDA tests then measure backend consistency.
The complete smoke suite reports 171 PASS, 0 FAIL, 4 WARN, and 0 SKIP. All
warnings fall inside a predeclared tolerance for floating-point accumulation
order in factored CUDA reductions. The artifact's core, sensitivity, and
long-horizon evaluations comprise 108 TensorBoard runs and 242.1 logged
GPU-hours on a single NVIDIA GeForce RTX 3090 Ti.
Appendix~\ref{app:implementation_detail} records thresholds, warning
magnitudes, software versions, and the exact training pipeline.

%% file: sections/conclusion.tex
\section{Conclusion}
\label{sec:conclusion}

Optimizer-state topology determines which quantization errors matter. Policies
designed for Adam's two dense moments do not transfer unchanged to factored,
confidence, and projected states. Adaptive Log-Space quantization addresses the
representation layer with a block-adaptive nonzero interval and an exact-zero
code; state-specific precision, semantic grouping, and parameter protection
address how that representation enters each optimizer update.

Controlled experiments show that adaptive ranges reduce update error and EMA
drift, while exact-zero reservation preserves dormant state entries. A matched
adaptive-log comparison further shows that lower state reconstruction error can
coexist with much larger update error after the consuming operator. End to end,
\ALviii{} with \UFviii{} outperforms the evaluated bitsandbytes (bnb) 8-bit AdamW
baseline in perplexity and throughput at essentially the same optimizer-state
storage. CAME matches reference-level final perplexity across three seeds by
assigning \ALxvi{} to its non-negative states, and quantized Adafactor closes
most of its 100K-step gap through semantic grouping and parameter protection.
On APOLLO's projected path,
\UFviii{} momentum with \ALviii{} second moments reduces measured optimizer
state from 2078.7 to 1694.7 MiB with a 0.56 PPL gap.

The central conclusion is constructive: choose the numerical representation,
block topology, and precision from the state semantics and the operator that
consumes the state. This turns optimizer quantization from a uniform storage
substitution into a design that remains effective after the optimizer itself
has changed the geometry of its memory.

%% file: sections/appendix.tex
\section{Additional Results}
\label{app:additional_detail}

The appendix exposes numeric results that support the main figures without
moving their visual narrative out of the paper. Controlled probes use the
fixed generators recorded in the executed analysis notebook.

\subsection{Trace-Calibrated State Sensitivity}
\label{app:trace_calibrated}

Table~\ref{tab:trace_calibrated_full} gives the complete values summarized in
Figure~\ref{fig:controlled_fidelity}a. Each 32,768-element synthetic state is
calibrated to the tensor-wise median and median block $\log_2$ range of the
corresponding step-10K CAME trace. These controlled one-step operator probes
measure conditional sensitivity under trace-matched numerical regimes. Values
beyond the plotting range are reported as censored observations.

\begin{center}
\begin{minipage}{\columnwidth}
\captionsetup{type=table}
\captionof{table}{Trace-calibrated single-step update error (\%, lower is
better). bnb 8-bit false-zero counts report positive state entries reconstructed
as zero out of 32,768 elements.}
\label{tab:trace_calibrated_full}
\centering
\small
\resizebox{\linewidth}{!}{%
\begin{tabular}{llrrrr}
\toprule
State & Tensor role & \ALviii{} & \ALxvi{} & bnb 8-bit & False zeros \\
\midrule
$V_r$ & Attention & 0.154 & 0.001 & 0.434 & 0 \\
$V_r$ & MLP       & 0.652 & 0.003 & 4.361 & 0 \\
$V_r$ & Embedding & 1.069 & 0.006 & $>100$ & 26,857 \\
\midrule
$C_r$ & Attention & 0.034 & $<0.001$ & 0.133 & 0 \\
$C_r$ & MLP       & 0.320 & 0.001 & 0.745 & 0 \\
$C_r$ & LM head   & 0.470 & 0.002 & 1.402 & 0 \\
$C_r$ & Embedding & 1.162 & 0.006 & $>100$ & 15,461 \\
\bottomrule
\end{tabular}}
\end{minipage}
\end{center}

The controlled probe shows two mechanisms directly. Required precision depends
jointly on the traced range and the operator that consumes the state, and the
two bnb 8-bit embedding failures coincide with thousands of positive entries being
reconstructed as zero. These conditional probes isolate update sensitivity;
the end-to-end experiments aggregate the consequence across states, tensors,
and training steps.

\subsection{Sensitivity Tables}
\label{app:sensitivity_tables}

Table~\ref{tab:sensitivity_full} reports trajectory mean absolute error
relative to the path-specific full-precision reference in the 10K learning-rate
and batch-size sweeps. A collapsed run is recorded at its observed failure step
without a finite error assignment.

\begin{center}
\begin{minipage}{\columnwidth}
\captionsetup{type=table}
\captionof{table}{10K-step sensitivity. Entries are training-trajectory MAE from the
path-specific full-precision reference. (a) Learning-rate multipliers of the
common $10^{-3}$ preset at batch size 4. (b) Batch size at the nominal
path-specific learning rate.}
\label{tab:sensitivity_full}
\centering
\small
\textbf{(a) Learning-rate sweep}\par\smallskip
\resizebox{\linewidth}{!}{%
\begin{tabular}{lrrr}
\toprule
Configuration & $0.1\times$ & $1\times$ & $10\times$ \\
\midrule
AdamW, bnb 8-bit & 0.0098 & 0.0777 & collapse at 2.9K \\
AdamW, \UFviii{}+\ALviii{} & 0.0078 & 0.0890 & 0.2687 \\
CAME, \UFviii{}+\ALxvi{} $V,C$ & 0.0143 & 0.1499 & 0.2429 \\
CAME, \UFviii{}+\ALviii{} $V,C$ & 0.0495 & 0.1091 & collapse at 4.96K \\
CAME, \UFviii{}+\ALviii{} $V$ + FP32 $C$ & 0.0298 & 0.2811 & 0.2794 \\
Adafactor, \ALviii{} $B=2048$ & 0.0253 & 0.0266 & 0.0298 \\
APOLLO, \UFviii{}+\ALviii{} & 0.0119 & 0.0157 & 0.1464 \\
\bottomrule
\end{tabular}}
\medskip
\textbf{(b) Batch-size sweep}\par\smallskip
\resizebox{\linewidth}{!}{%
\begin{tabular}{lrrr}
\toprule
Configuration & Batch 4 & Batch 8 & Batch 16 \\
\midrule
AdamW, bnb 8-bit & 0.0098 & 0.0118 & 0.0111 \\
AdamW, \UFviii{}+\ALviii{} & 0.0078 & 0.0053 & 0.0050 \\
CAME, \UFviii{}+\ALviii{} $V,C$ & 0.0495 & 0.0603 & 0.0572 \\
Adafactor, \ALviii{} $B=2048$ & 0.0266 & 0.0249 & 0.0283 \\
APOLLO, \UFviii{}+\ALviii{} & 0.0119 & 0.0138 & 0.0148 \\
\bottomrule
\end{tabular}}
\end{minipage}
\end{center}

The sensitivity tables show that the AdamW ordering persists across the tested
batch sizes, while high-learning-rate behavior and CAME precision remain
path-specific.

\subsection{Artifact and Source Map}
\label{app:artifact_map}

The public implementation anchor is Adafactor8Bit v0.4.3, commit
\href{https://github.com/yanfeiwong/adafactor-8bit/commit/4f544d8b8eebaf50053a4e8a27096e79b049b480}{\texttt{4f544d8}}.
That commit contains the optimizer implementation used for the experiments,
not this manuscript or its figure-generation sources.

The public reproducibility artifact at
\href{https://github.com/yanfeiwong/al-quantization}{\texttt{yanfeiwong/al-quantization}}
uses the following claim-to-source map:
\begin{itemize}
    \item Figures~\ref{fig:state_landscape} and~\ref{fig:al_pipeline} are
    generated by the matching scripts under \path{scripts/paper_figures/} and
    are exported as PDF, SVG, and PNG.
    \item Figure~\ref{fig:state_heterogeneity} uses
    \path{fig03_trace_components.csv}, a tidy export from the raw state traces
    produced by \path{export_fig03_trace_data.py}; artifact filenames retain
    their original numeric stems independently of manuscript figure order.
    \item Figure~\ref{fig:controlled_fidelity} uses three checked CSV tables
    exported from executed notebook cells F7, A1, and D4.
    \item Table~\ref{tab:qema_probe} uses Section H of
    \path{theory_and_ablation_final.ipynb}; the notebook pins the transcribed
    SOLO source commit and records the complete matched/default-block results.
    \item Figure~\ref{fig:training_curves} and the benchmark tables use
    \path{tb_analysis_report.md}, generated from the TensorBoard events by
    \path{analyze_tb.py}.
    \item Reference/CUDA validation and environment claims use the paired
    Markdown/JSON smoke-test and environment reports generated by their
    reproducibility scripts.
\end{itemize}

Raw TensorBoard events and state-trace snapshots are included as ancillary
repository artifacts and excluded from the LaTeX submission inputs. The machine-readable
reports retain the values used by the manuscript. Windows peak memory is
reported from PyTorch allocation counters and optimizer-state storage from live
CUDA tensors; neither quantity should be interpreted as an operating-system
measurement of total resident VRAM.

\balance
\subsection{Implementation Validation and Environment}
\label{app:implementation_detail}

Reference Algorithm Alignment compares the unquantized Python path with
PyTorch AdamW and RMSprop, Hugging Face Adafactor, official CAME, and official
APOLLO at a parameter-difference tolerance of $10^{-5}$. CUDA Numerical
Consistency uses a predeclared policy: PASS below $10^{-5}$, WARN from
$10^{-5}$ to below $5\times10^{-3}$, and FAIL at or above
$5\times10^{-3}$. The current complete smoke run reports 171 PASS, 0 FAIL,
4 WARN, and 0 SKIP. The Python CAME mixed-shape comparison reaches
$3.26\times10^{-6}$ maximum parameter difference. Its CUDA counterpart reaches
$1.24\times10^{-5}$; all-\ALviii{} CUDA--Python CAME comparisons are around
$1.25\times10^{-4}$. These warnings arise in factored \texttt{atomicAdd}
reductions, where parallel accumulation order differs; \ALxvi{} and FP32
confidence variants remain below the strict threshold.

All reported runs use WikiText-103 raw data shuffled through a 10,000-example
streaming buffer with empty records removed, sequence length 512, dynamic
padding to a multiple of eight, causal-language-model collation, gradient
checkpointing, BF16 computation, and FP32 parameters. The learning rate is
constant after 1000 warmup steps, and no checkpoint is selected by validation
score. The 20K core and 10K sensitivity runs use TinyLlama-1.1B; the 100K
horizon uses GPT-2 124M.

The captured environment is Windows 11 build 26200, Python 3.13.9, PyTorch
2.12.1 for CUDA 13.2, cuDNN 9.2, NVIDIA driver 610.62, and MSVC 19.44.35223.
Result-relevant packages include Transformers 5.13.0, Datasets 4.4.1,
Accelerate 1.14.0, bitsandbytes 0.50.0.dev0, came-pytorch 0.1.3, and
apollo-torch 1.0.3. The artifact retains wheel hashes and the Markdown/JSON
environment and smoke-test reports without machine-local paths.

%% file: references.bib
@inproceedings{adam,
  title={Adam: A Method for Stochastic Optimization},
  author={Kingma, Diederik P. and Ba, Jimmy},
  booktitle={International Conference on Learning Representations},
  year={2015}
}

@inproceedings{adamw,
  title={Decoupled Weight Decay Regularization},
  author={Loshchilov, Ilya and Hutter, Frank},
  booktitle={International Conference on Learning Representations},
  year={2019}
}

@inproceedings{adafactor,
  title={Adafactor: Adaptive Learning Rates with Sublinear Memory Cost},
  author={Shazeer, Noam and Stern, Mitchell},
  booktitle={International Conference on Machine Learning},
  year={2018}
}

@inproceedings{dettmers8bit,
  title={8-bit Optimizers via Block-wise Quantization},
  author={Dettmers, Tim and Lewis, Mike and Shleifer, Sam and Zettlemoyer, Luke},
  booktitle={International Conference on Learning Representations},
  year={2022}
}

@inproceedings{fourbit,
  title={Memory Efficient Optimizers with 4-bit States},
  author={Li, Bingrui and Chen, Jianfei and Zhu, Jun},
  booktitle={Advances in Neural Information Processing Systems},
  volume={36},
  year={2023}
}

@inproceedings{came,
  title={{CAME}: Confidence-Guided Adaptive Memory Efficient Optimization},
  author={Luo, Yang and Ren, Xiaozhe and Zheng, Zangwei and Jiang, Zhuo and Jiang, Xin and You, Yang},
  booktitle={Proceedings of the 61st Annual Meeting of the Association for Computational Linguistics (Volume 1: Long Papers)},
  pages={4442--4453},
  address={Toronto, Canada},
  publisher={Association for Computational Linguistics},
  doi={10.18653/v1/2023.acl-long.243},
  url={https://aclanthology.org/2023.acl-long.243/},
  year={2023}
}

@inproceedings{apollo,
  title={{APOLLO}: {SGD}-like Memory, {AdamW}-level Performance},
  author={Zhu, Hanqing and Zhang, Zhenyu and Cong, Wenyan and Liu, Xi and Park, Sem and Chandra, Vikas and Long, Bo and Pan, David Z. and Wang, Zhangyang and Lee, Jinwon},
  booktitle={Proceedings of Machine Learning and Systems},
  volume={7},
  url={https://proceedings.mlsys.org/paper_files/paper/2025/hash/437bc4ccafd3fc6d4289bd10940be42b-Abstract-Conference.html},
  year={2025}
}

@inproceedings{galore,
  title={GaLore: Memory-Efficient LLM Training by Gradient Low-Rank Projection},
  author={Zhao, Jiawei and Zhang, Zhenyu and Chen, Beidi and Wang, Zhangyang and Anandkumar, Anima and Tian, Yuandong},
  booktitle={International Conference on Machine Learning},
  year={2024}
}

@article{coat,
  title={COAT: Compressing Optimizer States and Activation for Memory-Efficient FP8 Training},
  author={Xi, Haocheng and Cai, Han and Zhu, Ligeng and Lu, Yao and Keutzer, Kurt and Chen, Jianfei and Han, Song},
  journal={arXiv preprint arXiv:2410.19313},
  year={2024}
}

@article{solo,
  title={Pushing the Limits of Low-Bit Optimizers: A Focus on EMA Dynamics},
  author={Xu, Cong and Liang, Wenbin and Yu, Mo and Liu, Anan and Zhang, Ke-Yue and Wang, Shunli and Ma, Lizhuang and Wang, Jianyong and Wang, Jun and Zhang, Wei},
  journal={arXiv preprint arXiv:2505.00347},
  eprint={2505.00347},
  archivePrefix={arXiv},
  primaryClass={cs.LG},
  url={https://arxiv.org/abs/2505.00347},
  year={2025}
}

@article{quantstaleness,
  title={Understanding Quantization of Optimizer States in LLM Pre-training: Dynamics of State Staleness and Effectiveness of State Resets},
  author={Topollai, Kristi and Choromanska, Anna},
  journal={arXiv preprint arXiv:2603.16731},
  eprint={2603.16731},
  archivePrefix={arXiv},
  primaryClass={cs.LG},
  url={https://arxiv.org/abs/2603.16731},
  year={2026}
}

@article{stquant,
  title={{STQuant}: Spatio-Temporal Adaptive Framework for Optimizer Quantization in Large Multimodal Model Training},
  author={Liu, Minglu and Hu, Cunchen and Xu, Liangliang and Tang, Fengming and Wang, Ruijia and Yu, Fu},
  journal={arXiv preprint arXiv:2604.06836},
  eprint={2604.06836},
  archivePrefix={arXiv},
  primaryClass={cs.LG},
  url={https://arxiv.org/abs/2604.06836},
  year={2026}
}

@article{tinyllama,
  title={TinyLlama: An Open-Source Small Language Model},
  author={Zhang, Peiyuan and Zeng, Guangtao and Wang, Tianduo and Lu, Wei},
  journal={arXiv preprint arXiv:2401.02385},
  year={2024}
}

@techreport{gpt2,
  title={Language Models are Unsupervised Multitask Learners},
  author={Radford, Alec and Wu, Jeffrey and Child, Rewon and Luan, David and Amodei, Dario and Sutskever, Ilya},
  institution={OpenAI},
  year={2019}
}

@inproceedings{wikitext,
  title={Pointer Sentinel Mixture Models},
  author={Merity, Stephen and Xiong, Caiming and Bradbury, James and Socher, Richard},
  booktitle={International Conference on Learning Representations},
  year={2017}
}
